\documentclass[a4paper, 10pt, conference]{ieeeconf}      

\IEEEoverridecommandlockouts                              

\makeatletter
\let\MYcaption\@makecaption
\makeatother

\usepackage[font=footnotesize]{subcaption}

\makeatletter
\let\@makecaption\MYcaption
\makeatother

\usepackage{subcaption}

\title{\LARGE \bf
Towards Local Minima-free Robotic Navigation: Model Predictive Path Integral Control via Repulsive Potential Augmentation
}

\author{Takahiro Fuke$^{1}$, Masafumi Endo$^{1}$, Kohei Honda$^{2}$ and Genya Ishigami$^{1}$
\thanks{*This work was not supported by any organization}
\thanks{$^{1}$T. Fuke, M. Endo, and G. Ishigami are with the Space Robotics Group, Department of Mechanical Engineering, Keio University, Yokohama 223-8522, Japan
        {\tt\small 
        \{taka162uke, masafumi.endo\}@keio.jp,
        ishigami@mech.keio.ac.jp}
        }
\thanks{$^{2}$K. Honda is with the Mobility System Group, Department of Mechanical Systems Engineering, Nagoya University, Nagoya 464-8603, Japan 
        {
        \tt\small
        honda.kohei.f4@a.mail.nagoya-u.ac.jp
        }
        }%
}

\usepackage{amsmath}
\usepackage{amssymb}
 
\usepackage{amsthm}

\newtheorem*{property1}{Property 1}
\newtheorem*{property3}{Property 3}
\newtheorem*{property4}{Property 4}
\newtheorem*{definitiongm*}{Definition of global minimum}
\newtheorem*{definitionlm*}{Definition of local minimum}
\usepackage{comment}
\usepackage{bbm}
\usepackage{float}
\usepackage{siunitx}
\usepackage{multirow}
\usepackage{booktabs}
\usepackage{graphicx}
\usepackage{subcaption}

\begin{document}

\twocolumn[
\noindent
© 2024 IEEE. Personal use of this material is permitted. Permission from IEEE must be obtained for all other uses, in any current or future media, including reprinting/republishing this material for advertising or promotional purposes, creating new collective works, for resale or redistribution to servers or lists, or reuse of any copyrighted component of this work in other works.\\

\noindent
\textbf{Submitted article:}\\
T. Fuke, M. Endo, K. Honda, and G. Ishigami ``Towards Local Minima-free Robotic Navigation: Model Predictive Path Integral Control via Repulsive Potential Augmentation,'' \textit{Under review for IEEE/SICE International Symposium on System Integration}, 2025.
]
\thispagestyle{empty}
\pagenumbering{gobble}
\clearpage

\maketitle
\thispagestyle{empty}
\pagestyle{empty}

\begin{abstract}
Model-based control is a crucial component of robotic navigation.
However, it often struggles with entrapment in local minima due to its inherent nature as a finite, myopic optimization procedure.
Previous studies have addressed this issue but sacrificed either solution quality due to their reactive nature or computational efficiency in generating explicit paths for proactive guidance.
To this end, we propose a motion planning method that proactively avoids local minima without any guidance from global paths.
The key idea is \emph{repulsive potential augmentation}, integrating high-level directional information into the Model Predictive Path Integral control as a single repulsive term through an artificial potential field.
We evaluate our method through theoretical analysis and simulations in environments with obstacles that induce local minima.
Results show that our method guarantees the avoidance of local minima and outperforms existing methods in terms of global optimality without decreasing computational efficiency.
\end{abstract}


\begin{figure}[t]
  \centering
  \captionsetup[subfigure]{justification=centering}
  \begin{subfigure}[t]{0.95\columnwidth}
    \centering
    \includegraphics[width=\textwidth]{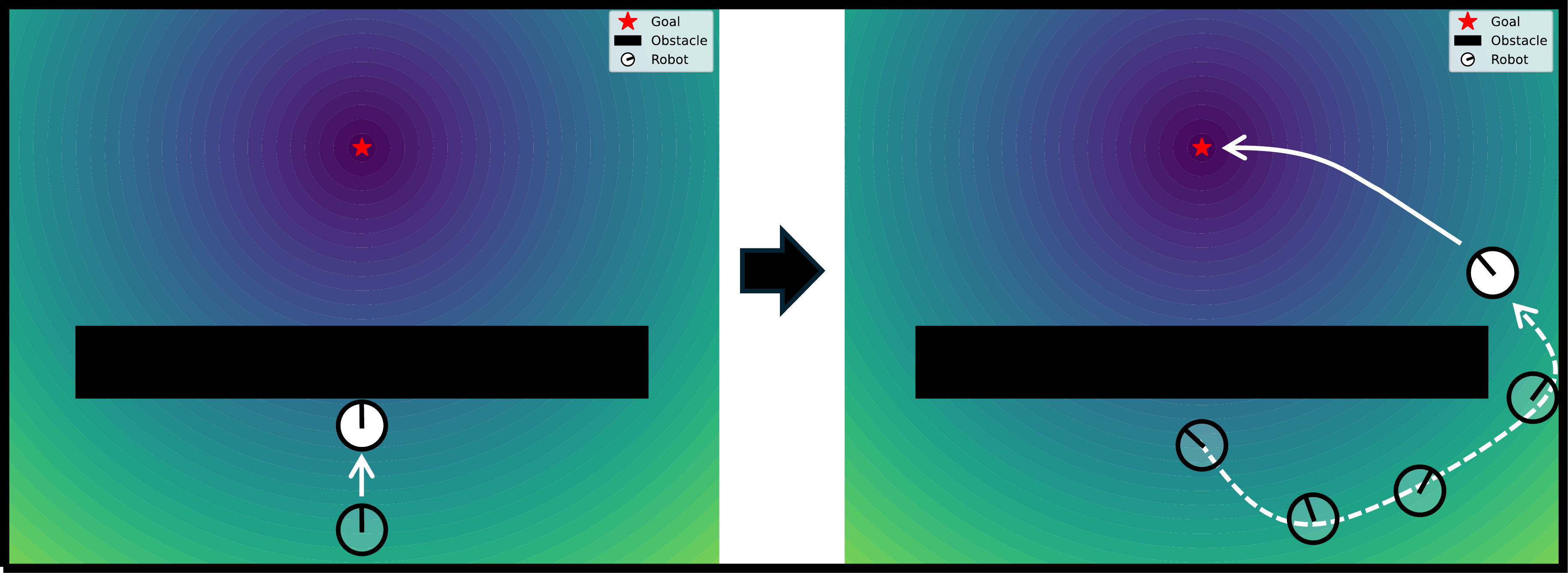}
    \caption{Reactive methods: Myopic, suboptimal solutions}
    \label{fig:intro_three_images_a}
  \end{subfigure}
  
  \vspace{0.5em}
  
  \begin{subfigure}[t]{0.95\columnwidth}
    \centering
    \includegraphics[width=\textwidth]{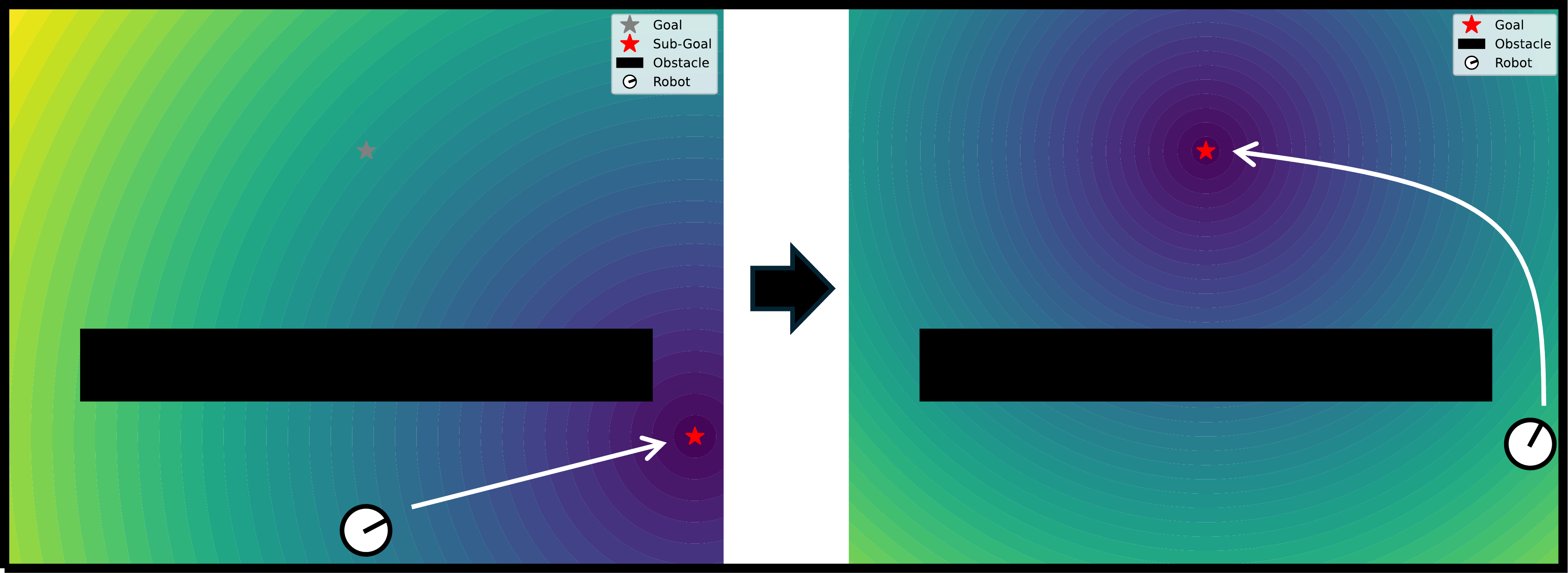}
    \caption{Proactive methods: Better optimality, requiring path search efforts}
    \label{fig:intro_three_images_b}
  \end{subfigure}
  
  \vspace{0.5em}
  
  \begin{subfigure}[t]{0.95\columnwidth}
    \centering
    \includegraphics[width=\textwidth]{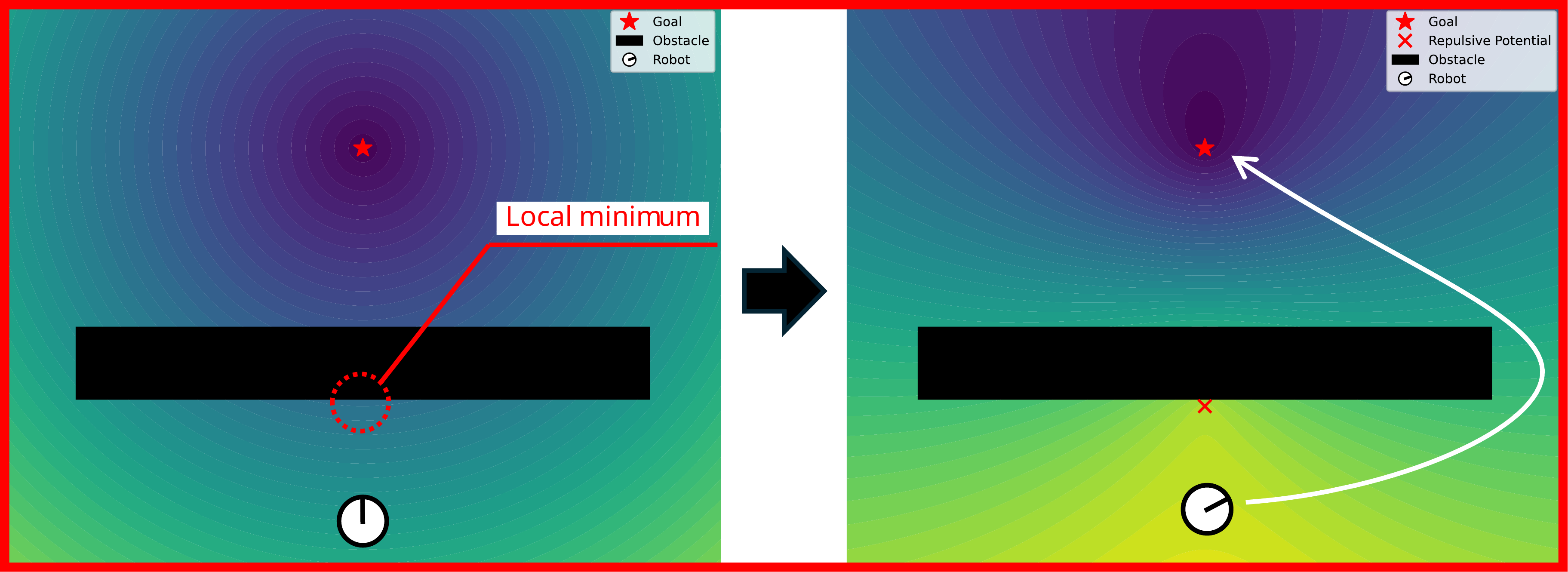}
    \caption{Ours: Near-optimal solutions by proactive view via APF\\ with local optimization, without additional computational effort}
    \label{fig:intro_three_images_c}
  \end{subfigure}

  \caption{Illustrative comparison of motion planning methods, motivated by the local minima problem around a large obstacle. Methods vary in solution optimality and computational efficiency, as detailed in subcaptions (a)-(c).}
  \label{fig:intro_three_images}
\end{figure}

\section{Introduction}\label{introduction}
{
Model Predictive Path Integral (MPPI) control~\cite{8558663}, a variant of sampling-based Model Predictive Control (MPC)~\cite{KAZIM2024100931}, is a powerful tool for robotic navigation that enables high-frequency optimal control by solving finite-horizon optimization problems.
MPPI offers advantages over traditional MPC in handling non-convex and nonlinear problems owning to its gradient-free trajectory rollouts.
However, much like traditional MPC, MPPI's inherent focus on generating sequential solutions over finite time horizons often leads robots into \emph{local minima}---suboptimal positions that appear optimal within a local neighborhood.
This myopic approach can result in performance far from global optimality and, in the worst case, mission failure due to entrapment in local minima.
Achieving global optimality is thereby crucial to solving real-world problems, such as long-horizon navigation in unstructured environments.

Existing studies have explored two main strategies to secure global optimality in model predictive techniques: 1) enhancing the quality of solutions within the \emph{local optimization} framework, and 2) incorporating \emph{global guidance} through integration with path planners. The former strategy includes reactive planning, such as the recovery behavior in Nav2~\cite{macenski2020marathon2}, which can recognize and avoid encountered local minima, but often remains suboptimal due to its ad hoc nature. The latter strategy takes proactive planning over a long horizon by generating reference paths a priori as global guidance. Various path planners, including search- and sampling-based algorithms, have been employed for different use cases~\cite{9636306, 9709214, https://doi.org/10.1002/rob.21959}.
Nevertheless, this strategy requires additional computational costs for reference path search, preventing tight integration of distinct planning steps at high frequency.
Global path planners also disregard robot dynamics, posing challenges in translating planned paths into feasible local motions~\cite{kuwata2008motion, 9773043}.} Recent studies have attempted to address these challenges by using machine learning techniques instead of conventional path planners \cite{10449376}. Yet, these approaches often involve complex implementations, limiting their broader applicability.

To free from the curse of finite time horizons, we propose a motion planning method that bridges the gap between local optimization and global guidance to avoid local minima (Fig. \ref{fig:intro_three_images}).
The key idea is \emph{repulsive potential augmentation} (RPA), which integrates high-level directional information into the MPPI framework.
This integration introduces a single \emph{repulsive term} into the optimization objective using an artificial potential field~\cite{warren1989global}.
Such a soft global guidance improves global optimality without explicit path search, resulting in computationally efficient planning.
Our approach requires minimal modifications to existing optimization problems, making it practical and extensible to various model predictive techniques. However, we prefer MPPI due to its high compatibility with our approach, especially in terms of solving speed and stability, given its gradient-free nature.

We evaluate our method through theoretical analysis and simulations in environments with different sizes of obstacles, which induce potential local minima entrapment.
Our analysis guarantees its ability to avoid entrapment in local minima. 
Experimental results also confirm that our method outperforms existing strategies in terms of global optimality and computational efficiency.
\section{MPPI Review}\label{mppi_review}
{
    MPPI is an advanced sampling-based method for solving stochastic optimal control problems. It can handle nonlinear systems, non-convex cost functions, and constraints in complex motion planning tasks. This section outlines MPPI's fundamental principles and key features relevant to our proposal.
    \subsection{Basics of MPPI}\label{basics_of_mppi}
    {
        MPPI considers nonlinear dynamical systems with Gaussian noise added to the control input:
        \begin{equation}\label{mppi_stochastic_dynamical_system}
            {\bf{x}}_{t+1} = {\bf{F}}({\bf{x}}_{t},{\bf{u}}_t+\delta{\bf{u}}_t),
        \end{equation}
        where ${\bf{x}}_{t}\in\mathbb{R}^{n}$, ${\bf{u}}_{t}\in\mathbb{R}^{m}$, and $\delta{\bf{u}}_{t}\in\mathbb{R}^{m}$ represent the state, control input, and Gaussian control noise with zero mean and covariance matrix $\Sigma$ at time $t$, respectively. For a finite horizon $t\in\{0,1,\cdots,T\}$, the stochastic optimal control problem to find the optimal control sequence ${\mathbf{U}}^{*}=\{{\bf{u}}_{0}^{*},{\bf{u}}_{1}^{*},\cdots,{\bf{u}}_{T-1}^{*}\}$ can be formulated as:
        \begin{multline}
            {\mathbf{U}}^{*}=\underset{\mathbf{U}}{\operatorname{argmin}}\,
            \mathbb{E}\left[\phi({\bf{x}}_{T}) + \sum_{t = 0}^{T-1}\left(c({\bf{x}}_{t}) + \frac{\lambda}{2}{\bf{u}}_{t}^{\rm{T}}\Sigma^{-1}{\bf{u}}_{t}\right)\right], \\
            \label{mppi_stochastic_optimization_problem}
        \end{multline}
        subject to
        \begin{align}
            & {\bf{x}}_{t+1} = {\bf{F}}({\bf{x}}_{t}, {\bf{u}}_t + \delta{\bf{u}}_t),\, \delta {\bf{u}}_{t}\sim \mathcal{N}(0,\Sigma), \label{mppi_sop_dyn}& \\
            & {\bf{x}}_{0} = {\bf{x}}_{\rm{init}}. & \label{mppi_sop_init}
        \end{align}
        Here, $\phi(\cdot)$ and $c(\cdot)$ represent the arbitrary terminal and state-dependent running costs, respectively.
        The term following the state-dependent running cost is the control cost, expressed as a quadratic form of the control input. ${\bf{x}}_{\rm{init}}$ denotes the initial condition, which corresponds to the current state.
        
        MPPI solves the problem in \eqref{mppi_stochastic_optimization_problem} using the importance sampling technique.
        The method begins with an initial estimated control sequence ${\mathbf{\widehat{U}}}=\{\widehat{{\bf{u}}}_{0},\widehat{{\bf{u}}}_{1},\cdots,\widehat{{\bf{u}}}_{T-1}\}$. It then generates $K$ perturbed control sequences $\mathbf{U}_k$ by adding Gaussian noise $\mathcal{N}(0,\Sigma_{\epsilon})$ to $\mathbf{\widehat{U}}$, where $\Sigma_{\epsilon}$ is the covariance matrix for the sampling distribution. These perturbed sequences simulate $K$ system trajectories, or \emph{rollouts}. The costs associated with these rollouts are used to compute a weighted average, which forms the basis for updating the control sequence.
The cost of the $k$-th rollout is denoted as:
\begin{equation}
J (\mathbf{x}_\text{init}, \mathbf{U}_k) = \phi({\bf{x}}_{T,k}) + \sum_{t = 0}^{T-1}\left(c({\bf{x}}_{t,k}) + {\lambda}{\widehat{\bf{u}}_{t}}^{\rm{T}}\Sigma_{\epsilon}^{-1}{\bf{u}}_{t,k}\right), \label{rollout_cost}
\end{equation}
and the optimal control input ${\bf{u}}_{t}^{*}$ is updated as follows:
\begin{equation}\label{mppi_control_update}
    {\bf{u}}_{t}^{*}={\widehat{\bf{u}}_{t}}+\frac{\sum_{k = 0}^{K-1} \exp{\left(-(1/\lambda) (J (\mathbf{x}_\text{init}, \mathbf{U}_k)-\rho)\right)}\delta {\bf{u}}_{t,k}}{\sum_{k = 0}^{K-1} \exp{\left(-(1/\lambda) (J (\mathbf{x}_\text{init}, \mathbf{U}_k)-\rho)\right)}},
\end{equation}
where $\lambda \in {\mathbb{R}}^{+}$ is the hyperparameter called temperature parameter, and $\rho$ is the minimum cost to prevent numerical over or underflow. The updated solution in (\ref{mppi_control_update}) minimizes the Kullback-Leibler divergence with the optimal control distribution characterized by the Boltzmann distribution~\cite{8558663}. 
The rollouts in MPPI are independent one another and therefore, MPPI is applicable for GPU-based parallel computation. MPPI does not need gradient calculations and iterative solution updates. This flexibility enables the use of arbitrary prediction models and cost designs, making it a subject of recent interest~\cite{10161511, 10379139, 10258400, 10611227, 10606099}.
    }

    \subsection{State and Control Input Constraints of MPPI}\label{state_and_control_input_constarints_of_mppi} 
    {
        MPPI's sampling-based scheme handles constraints differently from traditional MPC. In motion planning, two primary constraints are control and state constraints. The control constraints are satisfied by clipping perturbed control sequences, ensuring sampled inputs remain within feasible range. The state constraints are addressed by imposing large penalties on rollouts violating these constraints. While conceptually similar to soft constraints in traditional MPC, large penalties in MPPI transform these into hard constraints, expressed with indicator functions.
    }
}
\section{Local Minima-Free Navigation}\label{local_minima_free_cost_formulation}
{
    \subsection{Task Statement}\label{task_statement}
    {
        Point-to-goal navigation using MPPI without global reference paths is challenging, as the robot may become trapped in \emph{local minima}---suboptimal positions that appear optimal within a local neighborhood. This problem is due to the cost design and constraints.
        To describe the local minima issue, we consider a typical scenario in which a robot tries to avoid a rectangular obstacle in a planar environment (Fig.~\ref{fig:theoretical analysis}). This obstacle represents the simplest convex shape with straight edges, which primarily causes local minima. 
        We define a coordinate system with an obstacle of width $W$ and height $H$, centered at $\mathbf{p}_{\text{obs}} = [0, H/2]^{\top}$. The goal is at $\mathbf{p}_{\text{goal}} = [0, y_{\text{goal}}]^{\top}$, where $y_{\text{goal}} > H$.
        
        While this coordinate system simplifies our analysis, the derived properties remain valid for any chosen coordinate system. Although we analyze this scenario in a planar environment, the configuration exhibits axial symmetry with respect to the y-axis.
        This symmetry extends our discussion and results to three-dimensional scenarios, where the obstacle could be a vertical wall or similar structure.

    }
    \subsection{Definition of Global Minimum and Local Minima}\label{definition_of_global_minimum_and_local_minima}
    {        
        MPPI is a zeroth-order optimization method that does not require gradient information of the cost function. We define \emph{global minimum} and \emph{local minima} in MPPI analogously to other zeroth-order methods considering the specific context of robotic navigation.
        
        In robotic navigation, the {global minimum} is an optimal position where no further change for control input is needed, while {local minima} are suboptimal positions that appear optimal within a local neighborhood.
        We formalize these concepts by $\mathbf{U}_{\text{const}} \in \mathbb{R}^{m \times T}$, a constant control sequence that forces the robot to stay at the current position, assumed to be position-independent for simplicity. In our planner environment configuration, this concept can be represented by the zero control sequence.
        With $\mathbf{U}_{\text{const}}$, we define {global minimum} and {local minimum} for MPPI as follows:
        \begin{definitiongm*}\label{approach_def_gm}
        A position $\mathbf{p}^*$ is a global minimum if it satisfies the following:
        \[J(\mathbf{p}^*, \mathbf{U}_{\text{\rm{const}}}) \leq J(\mathbf{p}, \mathbf{U}_{\text{\rm{const}}}), \; ^\forall \mathbf{p}.\]
        \end{definitiongm*}
        
        \begin{definitionlm*}\label{approach_def_lm}
        A position $\hat{\mathbf{p}}^*$ is a local minimum if it satisfies the following:
        \begin{enumerate}
            \item Local minimality: $^\exists \delta > 0$ such that
            \[J(\hat{\mathbf{p}}^*, \mathbf{U}_{\text{\rm{const}}}) \leq J(\mathbf{p}, \mathbf{U}_{\text{\rm{const}}}), \; ^\forall \mathbf{p} \text{ with } \|\hat{\mathbf{p}}^* - \mathbf{p}\| < \delta.\]
            \item Non-global minimality: 
            
            $ J(\mathbf{p}^*, \mathbf{U}_{\text{\rm{const}}}) < J(\hat{\mathbf{p}}^*, \mathbf{U}_{\text{\rm{const}}})$.
        \end{enumerate}
        \end{definitionlm*}
        MPPI employs Gaussian random noise to explore the solution space independently of the cost function, aiding in avoiding local minima. However, MPPI requires careful parameter tuning such as planning horizon and Gaussian variance, often increasing computation time or degrading solution quality.
        These challenges highlight the importance of designing local minima-free cost functions for MPPI-based local minima-free navigation.
    }
    \subsection{Conventional Cost Formulation}\label{conventional_cost_formulation}
    {
        In MPPI, the cost-to-go function $J$ includes a terminal cost $\phi(\cdot)$ and a state-dependent stage cost $c(\cdot)$ according to the task objectives. For point-to-goal navigation tasks, these costs are generally designed using the squared Euclidean distance to the target position:
        \begin{align}
            \phi_{\text{baseline}}(\mathbf{p}) &= 
            c_{\text{baseline}}(\mathbf{p}) \nonumber \\
            &= \|\mathbf{p}_{\text{goal}} - \mathbf{p}\|^2 +  w_{\text{obst}} \cdot \mathbbm{1}_{\text{obst}}(\mathbf{p}), \label{approach_conv_cost}
        \end{align}
        where $w_{\text{obst}} \in {\mathbb{R}}^{+}$ is the positive collision penalty coefficient. It works with the indicator function $\mathbbm{1}_{\text{obst}}(\mathbf{p})$ to represent the obstacle constraint. The indicator function is defined using a set of positions $A$ occupied by obstacles:
        \begin{equation}\label{approach_ind_func}
            \mathbbm{1}_{\text{obst}}(\mathbf{p}) = \begin{cases}
                1 & \text{if } \mathbf{p} \in A \\
                0 & \text{if } \mathbf{p} \notin A
            \end{cases}.
        \end{equation}
        
        Under the configuration described in Section \ref{task_statement}, the conventional cost formulation in (\ref{approach_conv_cost}) results in a {local minimum} at $\mathbf{p}_{\text{minimum}}=[0, 0]^{\top}$. This makes it challenging for the robot to reach the target position when motion planning starts anywhere from $y < 0$, as illustrated in Fig. \ref{fig:theoretical analysis}.
    }

    \subsection{Repulsive Potential-Augmented Cost Formulation}\label{proposed_local_minima_free_cost_formulation}
    {
        As discussed in the \ref{conventional_cost_formulation}, local minima are inevitable as long as the conventional cost formulation is applied. We thus propose a new \emph{repulsive potential-augmented cost} that incorporates two key modifications: the use of Euclidean distance instead of squared Euclidean distance, and the addition of a repulsive term to encourage the robot to move away from {local minimum}. Our proposed formulation is
        \begin{align}
            \phi_{\text{proposed}}(\mathbf{p}) 
            & = c_{\text{proposed}}(\mathbf{p}) \nonumber \\
            & = \|\mathbf{p}_{\text{goal}} - \mathbf{p}\| - \alpha\|\mathbf{p}_{\text{minimum}} - \mathbf{p}\| \nonumber\\
            & \quad + w_{\text{obst}} \cdot \mathbbm{1}_{\text{obst}}(\mathbf{p}), \label{approach_prop_cost}
        \end{align}
        where $\alpha \in (0, 1)$ balances the repulsion from {local minimum} and attraction to the target position.
        The repulsive potential-augmented cost function in \eqref{approach_prop_cost} is simple to implement, only adding a single term when a local minimum is detected, which offers advantages for system integration. 
    }
    \begin{figure}[t]
    \centering
    \includegraphics[width=0.96\columnwidth]{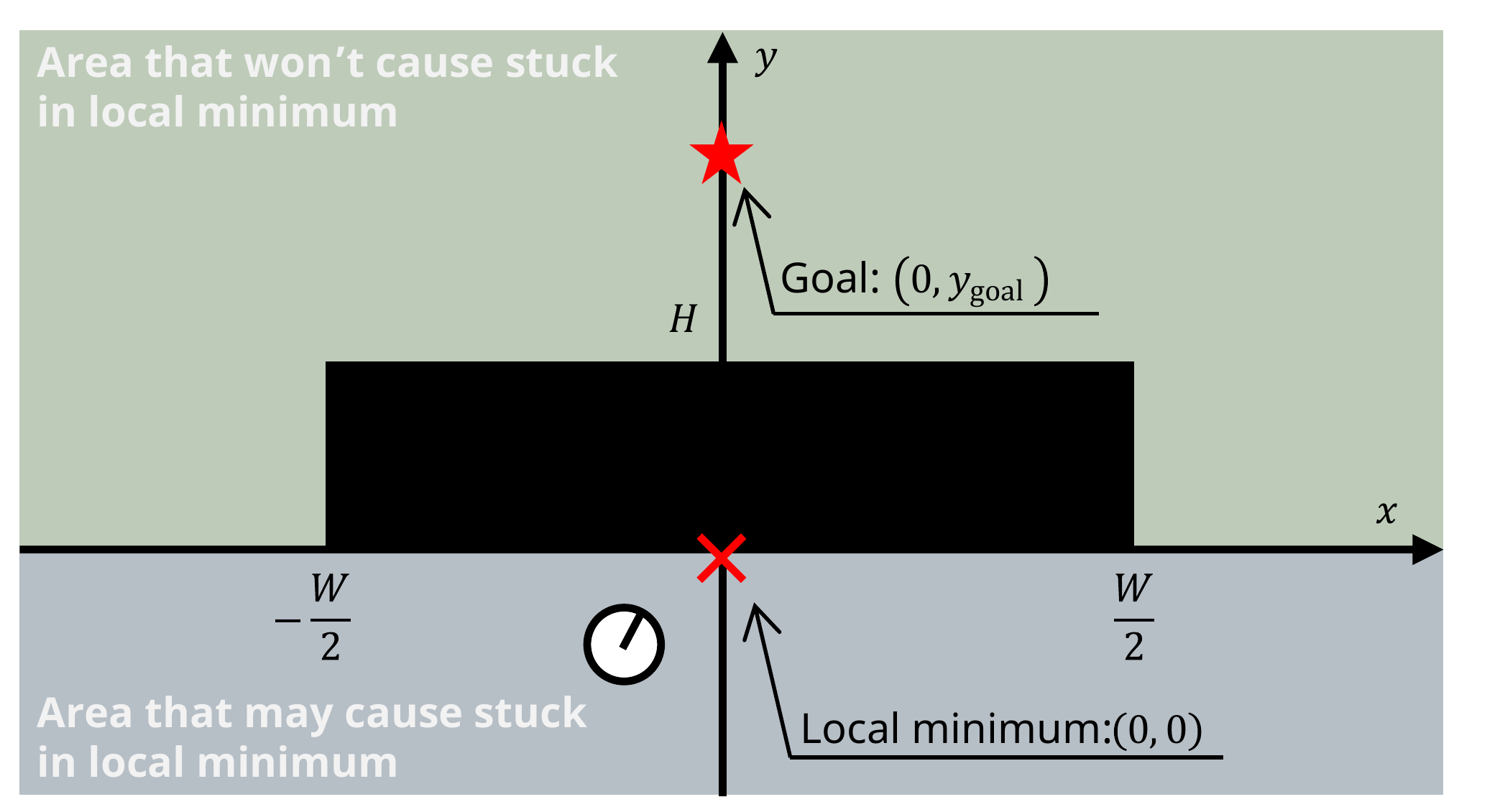}
    \caption{Coordinate system setting for theoretical analysis. The position indicated as {local minimum} and the associated regions illustrated are based on the conventional squared Euclidean distance-based cost. 
    }
    \label{fig:theoretical analysis}
    \end{figure}
    
    \subsection {Analysis of Repulsive Potential-Augmented Cost}\label{theoritical_analysis_proposed_local_minima_free_cost}
    {
        To demonstrate the cost described in (\ref{approach_prop_cost}) is free from local minima, we show that $\mathbf{p}_{\text{goal}}$ remains the global minimum and no other local minima exist. Based on (\ref{approach_prop_cost}), the cost-to-go $J_{\text{proposed}} (\mathbf{p}, \mathbf{U}_{\text{const}})$ is calculated by (\ref{rollout_cost}) as follows:
        \begin{align}
            & J_{\text{proposed}}(\cdot) = (T+1)(g(\mathbf{p})+w_{\text{obst}}\cdot\mathbbm{1}_{\text{obst}}(\mathbf{p})) + C_{\text{control}}, \label{eq:J_proposed_simplified} \\
            & g(\mathbf{p}) = \|\mathbf{p}_{\text{goal}} - \mathbf{p}\| - \alpha\|\mathbf{p}_{\text{minimum}} - \mathbf{p}\|. \label{eq:g_p_definition}
        \end{align}
        where $C_{\text{control}}$ is the constant term dependent on the constant control sequence, and $(T+1)$ is the horizon-dependent coefficient, both independent of position.
        
        We focus on $g(\mathbf{p})$, representing the cost function in the obstacle-free region to analyze the local minima-free property of $J_{\text{proposed}} (\mathbf{p}, \mathbf{U}_{\text{const}})$.
        Our analysis shows that $g(\mathbf{p})=g(x,y)$ satisfies the following four properties in the established coordinate system:
        \begin{enumerate}
        \item $\frac{\partial g(x,0)}{\partial x} < 0 \quad \text{for } 0 < x \leq \frac{W}{2}$
        \item $0 < \frac{\partial g(x,0)}{\partial x} \quad \text{for } -\frac{W}{2} \leq x < 0$
        \item $\frac{\partial g(x,y)}{\partial y} < 0 \quad \text{for } 0 \leq y \leq H$
        \item $g(0, y_{\text{goal}})$ is the unique global minimum, and there are no local minima.
        \end{enumerate}
        Property 1 to 3 collectively show that $J_{\text{proposed}} (\mathbf{p}, \mathbf{U}_{\text{const}})$ does not exhibit {local minima} along the perimeter of the obstacle. Property 4 establishes that $\mathbf{p}_{\text{goal}}$ is the unique {global minimum} and that no {local minima} exists in the obstacle-free region. Together, these properties indicate that the proposed cost formulation satisfies the criteria for a {local minima-free cost}. Detailed mathematical proofs for Property 1, 3, and 4 are in the appendix, while the proof for Property 2 follows analogously to that of property 1, with appropriate sign changes due to the reversed $x$ coordinate range.
    }
}
\section{Simulation Experiments}
{
    This section demonstrates the effectiveness of Repulsive Potential-Augmented MPPI (RPA-MPPI), integrating repulsive potential-augmented cost into the MPPI framework. We prepare rectangular obstacle scenarios, as this configuration aligns with our theoretical analysis.

    \subsection{Transition Model for 2D Navigation}
    {
        The robot is modeled as a simple point-mass non-holonomic system, represented as follows:
        \begin{equation}
            \begin{bmatrix}
                x_{t+1}\\
                y_{t+1}\\
                \theta_{t+1}
            \end{bmatrix}
        =
            \begin{bmatrix}
                x_{t}\\
                y_{t}\\
                \theta_{t}
            \end{bmatrix} 
        +
            \begin{bmatrix}
                v_{t}\cos{\theta_{t}}\\
                v_{t}\sin{\theta_{t}}\\
                \omega_{t}
            \end{bmatrix}
        \Delta t
        \label{eq:4},
        \end{equation}
        where $x_t$, $y_t$ represent the robot's position [m], and $\theta_{t}$ represents heading angle [rad] at the discrete time $t$. The input linear velocity $v_t$ is constrained to $[0, 1]$ m/s, and the input angular velocity $\omega_t$ is limited to $[-0.5, 0.5]$ rad/s. The time step $\Delta t$ is set to 0.1 s. This model is used for both the simulation state transition and the MPPI prediction model.
    }

    \subsection{Validation Settings}
    {
        Our validation process included three experimental scenarios, as shown in Fig. 3, characterized by varying obstacle widths: Short, Middle, and Long. For each scenario, goal tolerance was set to 1.0 m and all obstacles had a consistent 0.5 m safety margin. We tested 24 different initial robot states in each setup to ensure comprehensive evaluation.
        The comparative analysis included:
        (1) Standard MPPI (Std-MPPI) with a horizon of 50,
        (2) Std-MPPI with a horizon of 150,
        (3) Std-MPPI with a horizon of 50, guided by the A* algorithm~\cite{hart1968formal} as a global path planner (A*-MPPI), and
        (4) RPA-MPPI (ours) with a horizon of 50.
        The A* algorithm provides a resolution-optimal path when such a path exists. Therefore, A*-MPPI, which incorporates this algorithm, serves as a reliable baseline for comparison. They are implemented in PyTorch for GPU acceleration and run in the simulator~\cite{honda2023mppi}.
        
        Throughout the algorithm execution, the number of samples $K$ was set to 1000, covariance matrix $\Sigma_\epsilon$ to $\text{diag} \{1.00, 1.00\}$, and temperature parameter $\lambda$ to 0.10. For A*-MPPI, we used octile distance as the A* heuristic, set the grid size to 0.50 m, and used a 2.00 m look-ahead distance for determining subgoals on the reference path. For RPA-MPPI, taking account of the safety margin, $\mathbf{p}_{\text{minimum}}=[0, 9.0]^{\top}$ was assumed to be given in advance and set the repulsion coefficient to $\alpha = 0.75$, which provided a strong repulsive effect while maintaining sufficient goal attraction.
        The time limit is set as 50 seconds per navigation trial, and the following four metrics quantify algorithm performance:
        \begin{itemize}
            \item \textbf{Success Rate} (SR) [\%] evaluates robot safety based on the percentage of successful goal reaches without exceeding the time limit or colliding with obstacles.
            \item \textbf{Relative Difference in Success Time} (RDST) [\%] evaluates robot goal-reaching efficiency based on the average relative difference in success time compared to A*-MPPI for successfully completed scenarios.
            \item \textbf{Relative Difference in Path Length} (RDPL) [\%] evaluates path optimality based on the average relative difference in total traversed path length compared to A*-MPPI for successfully completed scenarios.
            \item \textbf{Computation Time} (CT) [s] evaluates computational efficiency based on average computational time for optimization per step.
        \end{itemize}
            }

    \begin{figure}
        \centering
        \includegraphics[width=0.95\columnwidth]{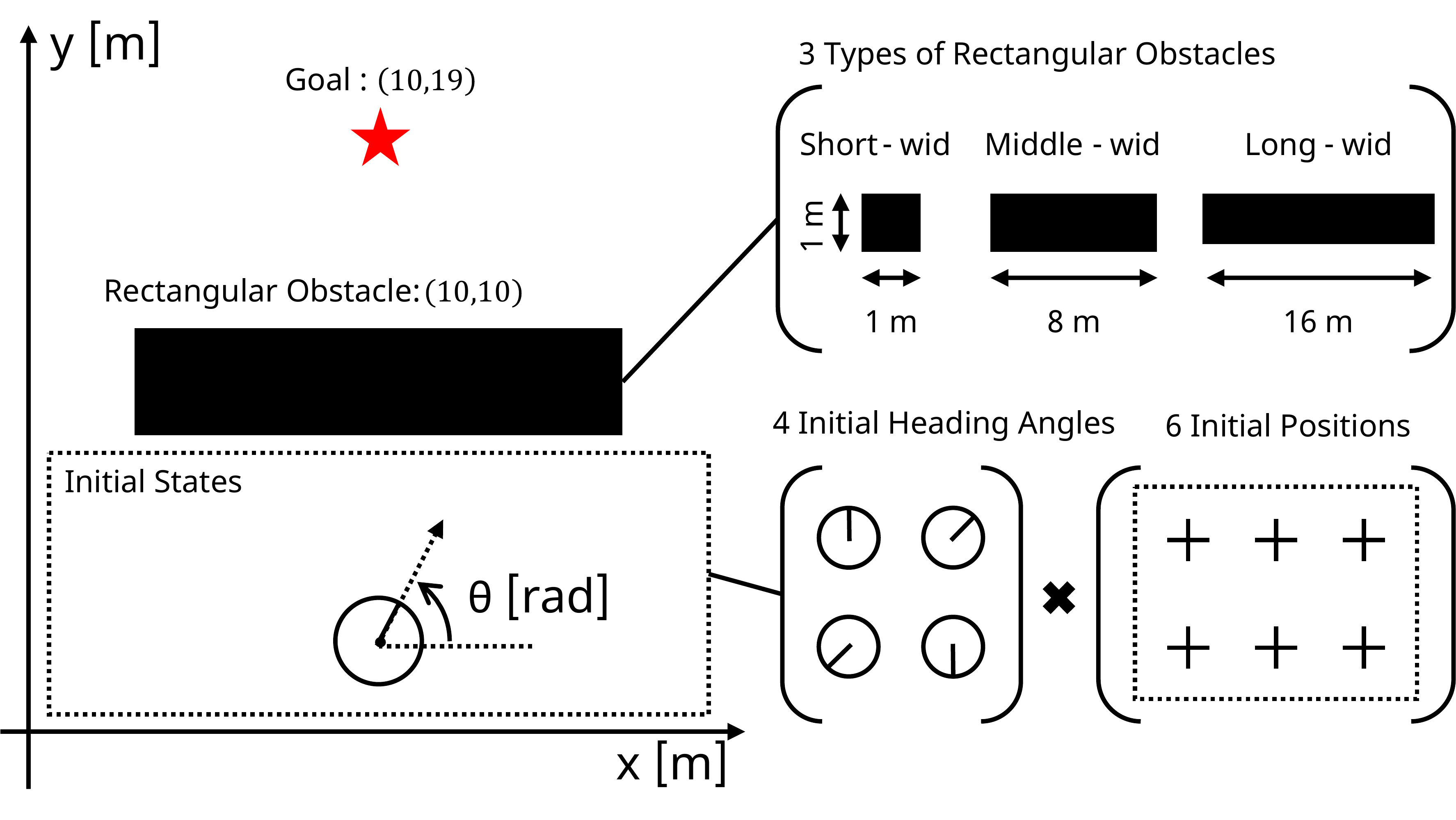}
        \caption{Three obstacle configurations are tested across 24 initial states. Initial Heading Angles [rad]: \{$\frac{\pi}{4}$, $\frac{\pi}{2}$, $\frac{5}{4}\pi$, $\frac{3}{2}\pi$\};  Initial positions [m]: \{(10, 1), (1, 1), (19, 1), (10, 8.5), (1, 8.5), (19, 8.5)\}}
        \label{fig:experimental settings}
    \end{figure}
\begin{table*}[t!]
\centering
\caption{Performance comparison of mppi-based navigation across varying obstacle configurations}
\begin{tabular}{lc|cccccccccccc}
\toprule
& & \multicolumn{4}{c}{Short-wid} & \multicolumn{4}{c}{Middle-wid} & \multicolumn{4}{c}{Long-wid}\\
\cmidrule(lr){1-2} \cmidrule(lr){3-6} \cmidrule(lr){7-10} \cmidrule(lr){11-14}
Planner Name & Horizon & SR$\uparrow$ & RDST$\downarrow$ & RDPL$\downarrow$ & CT$\downarrow$ & SR$\uparrow$ & RDST$\downarrow$ & RDPL$\downarrow$ & CT$\downarrow$ & SR$\uparrow$ & RDST$\downarrow$ & RDPL$\downarrow$ & CT$\downarrow$\\
\midrule \midrule
A*-MPPI & 50 & 100 & - & - & 0.248& 100 & - & - & 0.378& 100 & - & - & 0.572\\
Std-MPPI & 50 & 100 & \bf{-10.87} & \bf{+0.30} & \bf{0.044} & 67 & \bf{+4.27} & \bf{+0.80} & \bf{0.045} & 21 & \bf{-4.26} & \bf{+1.32} & \bf{0.045}\\
Std-MPPI & 150 & 100 & +25.18 & +9.92 & 0.131 & 92 & +31.05 & +8.31 & 0.131 & 42 & +39.27 & +6.97 & 0.133\\
\textbf{RPA-MPPI} & 50 & 100 & +7.54 & +23.16 & \bf{0.045} & \bf{96} & +9.29 & +21.73 & \bf{0.046} & \bf{88} & +10.11 & +23.11 & \bf{0.047}\\
\bottomrule
\end{tabular}
\label{tab:planner-comparison}
\end{table*}
\begin{figure}[t]
\centering
\includegraphics[width=0.95\columnwidth]{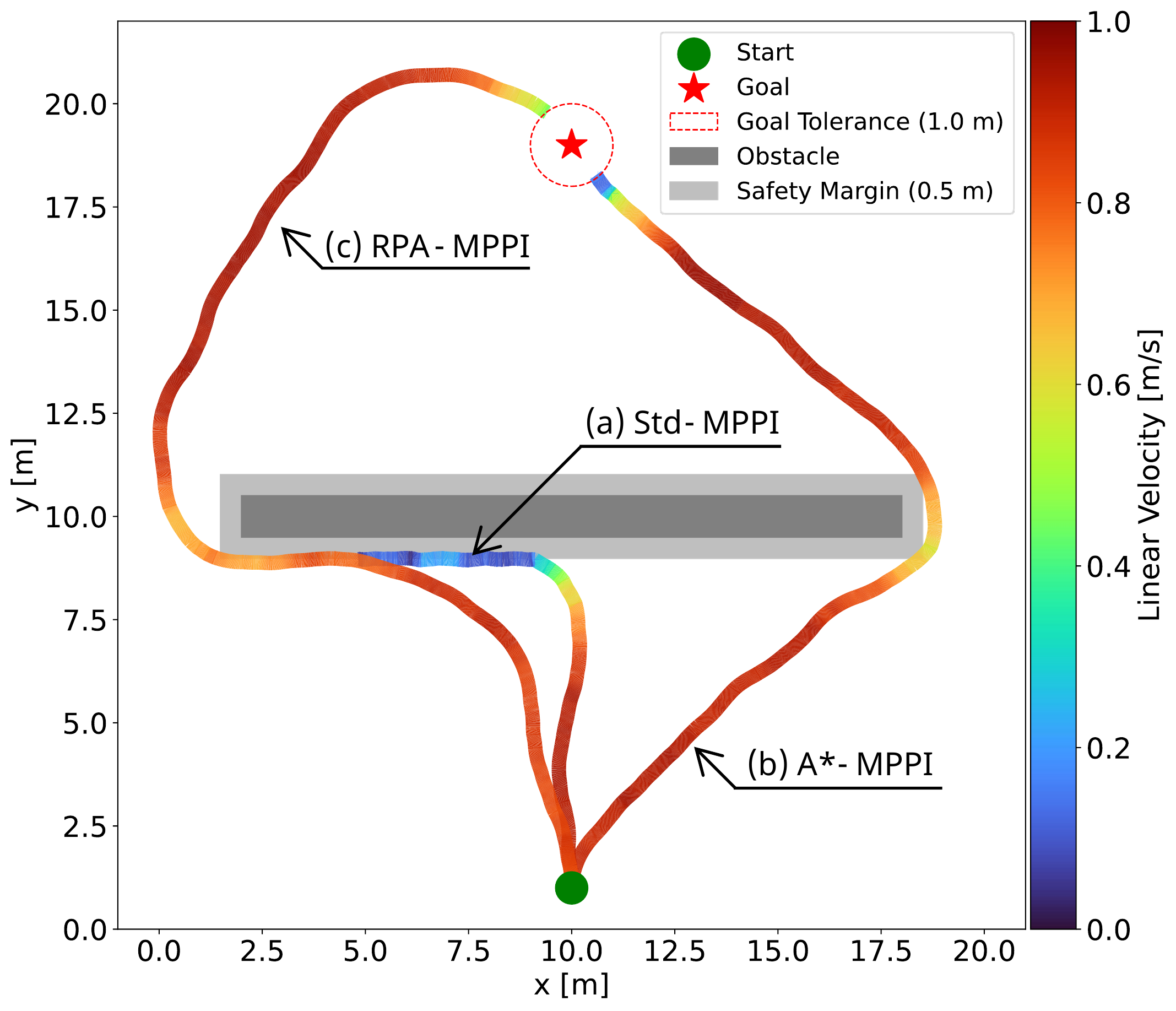}
\caption{Comparison of executed trajectories with Long-wid obstacle. (a) Std-MPPI: Fails to reach the goal due to local minima. (b) A*-MPPI: Reaches the goal with a reference path. (c) RPA-MPPI (ours): Reaches the goal without a reference path, adjusting speed and trajectory near obstacles.}
\label{fig:compare_trajectory}
\end{figure}
    \subsection{Results}\label{results}
    {
        Table \ref{tab:planner-comparison} summarizes the results across the three obstacle scenarios.
        RPA-MPPI outperforms Std-MPPI in SR across all scenarios. In the Long-wid scenario, longer horizons help Std-MPPI avoid local minima. However, this doesn't address the underlying issue noted in \ref{definition_of_global_minimum_and_local_minima}.
        A*-MPPI performs best in SR due to its explicit guidance; however, it also requires computational efforts across all scenarios.
        RPA-MPPI achieves comparable computational efficiency since it embeds proactive guidance steps in its optimization procedure without relying on additional path computations.
        It's worth noting that A*-MPPI requires a complete map prior to path planning. 
        This assumption is fragile, particularly in unstructured off-road environments, as recent studies have focused on robot navigation using only local motion planning~\cite{gasparion2022wayfast}.
        Our method is applicable even in an incomplete map as it incrementally identifies local minima. However, RPA-MPPI shows higher values in RDST compared to Std-MPPI, suggesting lower goal-reaching efficiency. RPA-MPPI also shows higher values in RDPL, indicating that the main factor for this inefficiency is the deviation from the shortest path.
        Fig. \ref{fig:compare_trajectory} provides qualitative insights into how different planners behave: RPA-MPPI successfully avoids local minima as the robot maintains its velocity throughout navigation. In contrast, Std-MPPI struggles to recover once trapped in local minima. Furthermore, comparing the paths traversed by RPA-MPPI and A*-MPPI reveals that the former deviates from the shortest path.
    }
    \subsection{Limitation and Possible Extensions}
    {
        \subsubsection{Dynamic Local Minima Detection}
        {
            As mentioned in \ref{results}, RPA-MPPI tends to traverse inefficient paths. This suggests that depending on the size of obstacles encountered, there may be situations where employing Std-MPPI is advantageous.
            Additionally, although we assume known {local minima} coordinates, it is rarely the case in real-world applications. We thus conducted additional experiments: RPA-MPPI with incorrectly identified local minima. As a result, it showed a failure in reaching the goal. 
            These results highlight the importance of accurate local minima detection and dynamic cost function adaptation in real-world applications. We assume such detection can be achieved by analyzing MPPI's output trajectories, potentially enabling {local minima} identification without global search and maintaining our method's performance advantage.
        }
        \subsubsection{Non-Convex Obstacle Handling}
        {
            Our method is optimized for convex obstacles, which presents challenges when dealing with non-convex shapes. While it may handle simple non-convex obstacles to some extent, it is likely to struggle with complex configurations such as U-shaped obstacles. Future work will focus on extending our approach to manage non-convex obstacles, possibly by decomposing or abstracting them into convex representations.
        }
    }
}

\section{Conclusion}
{
    We propose a motion planning method integrating a repulsive potential-augmented cost into the MPPI framework, called RPA-MPPI. This approach guides robots around obstacles while ensuring global convergence. Theoretical analysis and simulation experiments has validated its effectiveness, particularly in environments with wide convex obstacles where squared Euclidean distance-based cost often fails. Key challenges for real-world applications of RPA-MPPI include dynamic local minima detection and non-convex obstacle handling. Future work will focus on developing robust methods for these challenges, as well as extending the approach to 3D navigation scenarios.
}
\addtolength{\textheight}{-0cm}   

\section*{APPENDIX}
{
    This appendix provides proofs for the properties of $g({\bf{p}})=g(x,y)$ discussed in \ref{theoritical_analysis_proposed_local_minima_free_cost}. As a preliminary step, we present the partial derivatives of $g(x,y)$ for $[x,y]^{\top} \neq [0,0]^{\top}, [0,y_{\text{goal}}]^{\top}$:
        \begin{align}
            \frac{\partial g (x,y)}{\partial x} &= \frac{x}{\sqrt{x^2+(y_{\text{goal}}-y)^2}} - \frac{\alpha x}{\sqrt{x^2+y^2}}, \label{gx} \\
            \frac{\partial g (x,y)}{\partial y} &= \frac{-(y_{\text{goal}}-y)}{\sqrt{x^2+(y_{\text{goal}}-y)^2}} - \frac{\alpha y}{\sqrt{x^2+y^2}}. \label{gy}
        \end{align}

    \begin{property1}
        $\frac{\partial g(x,0)}{\partial x} < 0$ \rm{for} $0 < x \leq \frac{W}{2}$.
    \end{property1}
    \begin{proof}
    For $0 < x$, the inequality $\frac{\partial g (x,y)}{\partial x} < 0$ can be transformed as follows:
    \begin{align}
    & \frac{\partial g (x,y)}{\partial x} < 0 \nonumber \\
    & \iff \frac{1}{\sqrt{x^2+(y_{\text{goal}}-y)^2}}-\frac{\alpha}{\sqrt{x^2+y^2}} < 0 \nonumber \\
    & \iff x^2 < \frac{\alpha^2(y_{\text{goal}}-y)^2-y^2}{1-\alpha^2}. \label{prop2 fist eq}
    \end{align}
    Given that $y\leq0$, $0<y_{\text{goal}}$ and $0<\alpha<1$, the RHS of the final inequality in (\ref{prop2 fist eq}) is always positive. Therefore, it can be transformed as follows:
    \begin{equation}\label{prop2 eq2}
        0<x<\sqrt{\frac{\alpha^2(y_{\text{goal}}-y)^2-y^2}{1-\alpha^2}}.
    \end{equation}
    In particular, at $y=0$, equation \eqref{prop2 eq2} becomes
        \begin{equation}\label{prop2 eq3}
        0<x<y_{\text{goal}}\sqrt{\frac{\alpha^2}{1-\alpha^2}}.
    \end{equation}
    As $\alpha$ approaches 1, $y_{\text{goal}}\sqrt{\frac{\alpha^2}{1-\alpha^2}}$ grows without bound. By selecting an appropriate value for $\alpha$ sufficiently close to 1, this property can be theoretically satisfied for any given $W$.
    \end{proof}
    \vspace{-5mm}
    \begin{property3}
        $\frac{\partial g(x,y)}{\partial y} < 0$ \rm{for} $0 \leq y \leq H$.
    \end{property3}

    \begin{proof}
        Given that $H < y_{\text{goal}}$, the following inequality holds for $0 < y \leq H$:
        \begin{align}
            & \frac{\partial g (x,y)}{\partial y} = \frac{-(y_{\text{goal}}-y)}{\sqrt{x^2+(y_{\text{goal}}-y)^2}} - \frac{\alpha y}{\sqrt{x^2+y^2}} \nonumber \\
            &\leq \frac{-(y_{\text{goal}}-y)}{\sqrt{x^2+(y_{\text{goal}}-y)^2+y^2}}
            - \frac{\alpha y}{\sqrt{x^2+(y_{\text{goal}}-y)^2+y^2}}. \label{appendix_prop3_1}
        \end{align}
        Since $0<\alpha<1$, RHS of \eqref{appendix_prop3_1} is always negative. $\frac{\partial g (x,y)}{\partial y}$ is also smaller than or equal to this negative RHS. Therefore, $\frac{\partial g (x,y)}{\partial y}$ is always negative, and the property is satisfied.
    \end{proof}
    \vspace{-2mm}
    \begin{property4}
    $g(0, y_{\text{goal}})$ \rm{is the unique global minimum, and has no local minima.}
    \end{property4}
    
    \begin{proof}
        $g(x,y)$ is differentiable and continuous for $[x,y]^{\top} \neq [0,0]^{\top}, [0,y_{\text{goal}}]^{\top}$, and also continuous at $[x,y]^{\top} = [0,0]^{\top}$ and $[0,y_{\text{goal}}]^{\top}$ shown by the following limits:
        \begin{align}
            & \lim_{[x,y]\to [0,0]} g(x,y) = \lim_{r\to 0} g(r\cos{\psi}, r\sin{\psi}) \nonumber \\
            &\qquad \quad = \lim_{r\to 0} \left( \sqrt{(r\cos\psi)^2+(y_{\text{goal}}-r\sin\psi)^2} -\alpha r \right) \nonumber \\
            &\qquad \quad = y_{\text{goal}} = g(0,0), \label{g(0,0)} \\
            & \lim_{[x,y]\to [0,y_g]} g(x,y) = \lim_{r\to 0} g(r\cos{\psi}, r\sin{\psi} + y_{\text{goal}}) \nonumber \\
            &\qquad \quad= \lim_{r\to 0} \left( r - \alpha \sqrt{(r\cos\psi)^2 + (r\sin\psi + y_{\text{goal}})^2} \right) \nonumber \\
            &\qquad \quad= -\alpha y_{\text{goal}} = g(0, y_{\text{goal}}). \label{lim_to_ygoal}
        \end{align}

        Let $I= \{(x,y) \in \mathbb{R}^2| x^2+y^2 \leq R^2 \}$ be a bounded closed set for a sufficiently large positive real number $R$. By the Extreme Value Theorem, maximum and minimum values are guaranteed on this set. The following limit demonstrates that $g(x,y)$ grows unbounded as $r$ approaches infinity, ensuring that the minimum value exists within $x^2+y^2 < R^2$:
        \begin{align}
            & \lim_{r\to \infty} g(r\cos{\psi}, r\sin{\psi}) \nonumber \\
            & = \lim_{r\to \infty} \left( \sqrt{(r\cos\psi)^2 + (y_{\text{goal}} - r\sin\psi)^2} - \alpha r \right) \nonumber \\
            &= \lim_{r\to \infty} r \left( \sqrt{1 - \frac{2 y_{\text{goal}} \sin{\psi}}{r} + \frac{y_{\text{goal}}^2}{r^2}} - \alpha \right) = \infty. \label{g(0,y_g)}
        \end{align}
        Given $\alpha \in (0,1)$, there are no stationary points where both equations \eqref{gx} and \eqref{gy} equal zero simultaneously. This implies that differentiable points cannot be extrema, and consequently, cannot be the minimum. 
        Therefore, the minimum must occur at either $g(0, y_{\text{goal}})$ or $g(0,0)$. From equations \eqref{g(0,0)} and \eqref{g(0,y_g)}, we can deduce that $g(0, y_{\text{goal}}) < g(0,0)$, so $g(0,y_{\text{goal}})$ is the minimum. 
        Moreover, from Properties 1 and 2, we know that $g(0,0)$ is not even a local minimum. Hence, we can conclude that $g(0, y_{\text{goal}})$ is the unique global minimum, and there are no local minima.
    \end{proof}
}

\bibliographystyle{IEEEtran}
\bibliography{IEEEexample}
\end{document}